# ELEMENTOS DE INGENIERÍA DE EXPLOTACIÓN DE LA INFORMACIÓN APLICADOS A LA INVESTIGACIÓN TRIBUTARIA FISCAL[1]

**RODRIGO LOPEZ-PABLOS**[2]
rodrigo.lopezpablos@educ.ar

*Trabajo de investigación*

Esta versión:

Septiembre de 2013

## 1. INTRODUCCIÓN

Este trabajo intenta introducir a las ciencias económicas, las finanzas públicas y el área de la investigación tributaria fiscal en particular, elementos conceptuales, técnicas y herramientas provenientes de la literatura informático computacional, la inteligencia artificial y explotación de la información para que estas puedan contribuir al mejoramiento de las labores del investigador tributario en la administración fiscal.

Más allá de las ciencias económicas, los avances en el campo de la inteligencia artificial y los sistemas automáticos en computación representan una oportunidad como recurso técnico a ser adoptado por distintos ámbitos para el desarrollo humano. En este caso, las técnicas de ingeniería de explotación de la información así como han sido aplicadas interdisciplinariamente a distintas ramas del conocimiento [Britos, 2008], ofrecen una brecha potencial de aprovechamiento indubitable para las finanzas públicas si consideramos las mejoras recaudatorias y de transparencia fiscal que estas técnicas pueden significar para el ciudadano y la administración pública fiscal en general.

Una mejora en la función fiscal del Estado a través de un mejor control no solo repercutiría en un mayor acervo recaudatorio destinado a la satisfacción de la necesidad pública, con evidentes implicancias en el bienestar de las personas, la teoría del capital social afirma que mejores finanzas públicas fortalecidas a través de una ética fiscal que apuntale la función financiera del Estado, propiciará procesos virtuosos que resulten en un mayor bienestar general de las poblaciones administradas.

El apuntalamiento de la función financiera de los Estados no escapa a los planes de desarrollo de largo plazo, la agenda para el desarrollo posterior a los objetivos del milenio contempla, entre algunos de sus tópicos centrales, la necesidad de una revolución de transparencia en política fiscal y que esta se encuentre enfocada hacia

---

una mejor regulación financiera global, capaz de traducirse con el tiempo en una reducción en la corrupción, los flujos financieros ilícitos, el lavado de dinero, la evasión de impuestos, y el ocultamiento patrimonial de activos en la hacienda privada tanto como en el tráfico ilícito de drogas y armamentos [UN, 2013].

En ese mismo marco, la organización internacional de sociedades civiles (OISC) –la cual engloba a las organizaciones civiles de mayor importancia– en consonancia con lo anterior, considera pertinente que dentro de las transformaciones estructurales necesarias para mejorar el bienestar de las personas del planeta, es necesario reformar los sistemas financieros y fiscales de acuerdo con las obligaciones de derechos humanos para; entre otras cosas, garantizar mecanismos justos de renegociación de deuda soberana, redistribución de la riqueza y detener los flujos monetarios ilícitos.

Entre estos últimos objetivos pueden destacarse la necesidad de regulación de los mercados financieros, la abolición de los paraísos y guaridas fiscales, ya que estos en el sistema económico y financiero mundial actual favorecen y aumentan las desigualdades, obstaculiza la erradicación de la pobreza y la plena aplicación de todos los derechos humanos en todos sus niveles [CSO, 2013]. Frente a estos desafíos, las técnicas provistas por las ciencias informáticas y computacionales se presentan como instrumentos de oportunidad para afianzar estos nobles objetivos.

La necesidades de aprovechamiento tecnológico que surge desde las administraciones tributarias de los Estados nacionales fue advertido por varios autores en la órbita local [Milano, Steinberg; 2011, 2012]; sin embargo, las finanzas públicas no han sido gran objeto de difusión de dichas técnicas, dado que los procesos metodológicos de inteligencia artificial dirigidas a la explotación de la información han sido aplicadas casi exclusivamente por el sector privado, sin considerar menester las potencialidades que su aplicación pueda significar para la concreción de una administración pública más efectiva y transparente.

Mejoras en la eficacia del funcionamiento del Estado en general y de las administraciones públicas tributarias en particular, que al ser aplicadas con fines éticos[3] poseen el potencial latente de fortalecer el capital social de sus comunidades. Dentro de su modesto alcance, este trabajo tiene como objetivos básicos lo siguiente:

> **[i]:** Proveer al investigador tributario en administración pública no informático una variante para la investigación fiscal artesanal, de bajo costo y territorialmente focalizada.
> **[ii]:** Presentar una propuesta técnica para procesos de explotación de información para la investigación tributaria fiscal.
> **[iii]:** Introducir un conjunto de procedimientos que permitan la generación de conocimiento para bases de datos tributarios que favorezcan la transparencia fiscal.
> **[iv]:** Validar, a través de un caso de uso hipotético de investigación tributaria, un proceso de explotación de la información en la administración pública fiscal.

A continuación en la sección subsiguiente se procederá a la descripción del estado de la cuestión dentro del campo, continuando con los progresos en las finanzas públicas informáticas en la sección 3 para proseguir consecuentemente con el planteamiento de las hipótesis y el problema en la sección 4. La sección 5 plantea la propuesta metodológica en explotación de la información la que intenta validarse a través de un

---

[3] Lamentablemente, son bien conocidos los casos en que los propios Estados nacionales recurren a tecnologías informáticas con fines no éticos destinados al espionaje ilegal de personas físicas.

caso de uso hipotético en la sección 6. Las conclusiones son abordadas en la última sección.

## 2. EL ESTADO DE LA CUESTIÓN INFORMÁTICA

La información puede entenderse como un proceso cuantificado en bits, que a partir de un cúmulo de *datos* sin ninguna utilidad, esta puede evolucionar en distintos estadios, *información* y posteriormente en *conocimiento* como en sus estadios más desarrollados y de mayor potencialidad para la acción humana[4]. Este proceso puede describirse verticalmente de forma gráfica como se ilustra a continuación en la Figura 1.

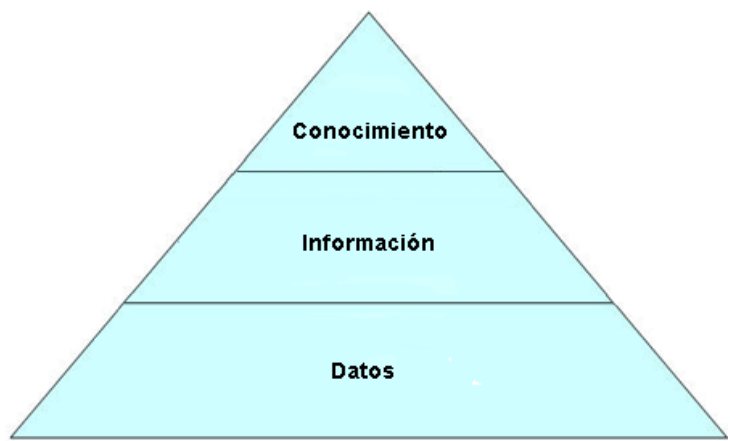

Fig. 1: Fases de la información

La pirámide de la información *ex supra* estampa las tres fases principales de la información en relación a la utilidad de la entidad que los gestiona ya sea un actor humano o computacional. Las técnicas de explotación de la información, dentro de los procesos de los procesos informáticos contribuyen a la transformación de los datos en conocimiento, fase última en la cual el insumo informativo ya posee una utilidad para el actor.

La información también pueden entenderse como un insumo invisible para la toma de decisiones tanto en economía como para cualquier actividad humana, incluso los individuos pueden postularse a sí mismos como recursos incrementales de información [Prada Madrid, 2008] puesto que todo sistema capaz de procesar información en el tiempo puede verse como una transformación de datos en conocimiento en función del propio actor que los procesa. Las técnicas y procesos de explotación de información como área de las ciencias informáticas, proporciona elementos fundamentales de un proceso más amplio que tiene como objetivo el descubrimiento de conocimiento en grandes bases de datos [Fayyad et al., 1996].

La explotación de la información toma los recursos informáticos disponibles, y a partir de la utilización de herramientas analíticas y de síntesis capaces de transformar la información en conocimiento, intenta generar conocimiento que contribuya a la toma de decisiones [Britos y Garcia-Martinez, 2009]. La ciencia computacional de la inteligencia artificial, y como subdisciplina, el campo de la explotación de la información, ha sido un área del conocimiento ampliamente usada por diversos sectores de la sociedad, en el cual propone un abordaje generalmente interdisciplinario.

---
[4] Varios autores consideran estadios aún más elevados incluyendo el de *idoneidad* y el de *sabiduría* como niveles informativos superiores al de *conocimiento*.

Como es sabido en el campo del arte, la explotación de información esta en sus primeros estadios [Garcia-Martinez *et al.*, 2013], es tan amplia en interdisciplina como las áreas plausibles de acción humana *per se*. Prueba de ello, son las múltiples dominios donde fueron aplicados, desde el descubrimiento de linfomas, identificación de acuerdos y desacuerdos entre grupos intrapartidarios y minorías interpartidarias en un Congreso Nacional, reconocimiento facial, clasificación de familias de asteroides, etc. [Britos, Garcia-Martinez *et al.*; *op. cit.*, 2011].

Hasta ahora fueron los sectores privados con acceso a técnicas avanzadas de tecnologías de la información los que han sabido aprovechar las virtudes de estas tecnologías para el logro de sus objetivos de mercado; tanto es así, que en su aplicación específica al sector privado ha dado lugar desde las ciencias informáticas a toda una subdisciplina denominada *Inteligencia de negocios*. La escasa literatura sobre la utilización de sistemas inteligentes de explotación de la información en el funcionamiento del Estado en administración pública con fines éticos son prueba de ello y ameritan la concreción de esta obra.

## 3. PROGRESOS EN EXPLOTACIÓN DE LA INFORMACIÓN SOBRE EL ÁREA FISCAL

Las administraciones tributarias de los Estados se hallan entre los más grandes productores, recolectores, consumidores y difusores de información de cada nación [Conde, 2000]. Los avances y la incidencia de la tecnología en el área fiscal supone la aparición de nuevos hechos imponibles así como el papel de los tributos en la nueva sociedad digital, marca un paradigma irreversible que la sociedad debe abordar desde el Estado, puesto que –entre otras cosas– las modalidades de intercambio de bienes y servicios presentan mayores dificultades a nivel fiscal cuando son efectuadas a través de la web [Berhouet, 2012]. Por esos motivos, no es casual que el plan estratégico 2011-2015 del máximo organismo de fiscalización del Estado argentino contemple la utilización de la información disponible para impulsar el cumplimiento de las leyes tributarias [AFIP, 2011].

La función de control es la metodología primordial utilizada por el organismo fiscalizador máximo del Estado argentino, hace al uso de herramental informático para el cruce de información, en la misma se cruza información desde distintas fuentes tales como datos del registros de la propiedad del inmueble, del automotor, registros de participación societaria, de la inspección general de justicia, etc. Para tales cruzamientos ya se advirtió la necesidad de crear matrices de riesgo a fin de eficientizar la tarea de fiscalización [Milano, *op. cit.*]; por otro lado, la alta complejidad de las transacciones exigieron la implantación de un no es casual que la AFIP invierta fuerte en tecnología informática de punta, el centro de cómputos central de la agencia federal tributaria es considerada un modelo de gestión en sí mismo, así como ha ganado varios premios internacionales por lo innovador del sistema de gestión implementado [Milano, Garello; *op. cit.*, 2010],

En uno de los trabajos más relevantes en el campo a la fecha, Steinberg [*op. cit.*] propuso un modelo de procesamiento de datos que permite definir y calificar un determinado grupo de grandes contribuyentes según su perfil de riesgo para el fisco. La metodología propuesta por Steinberg supone el armado de un *warehouse* a partir de distintas fuentes de data para, en segunda instancia, entablar un proceso de minería CRISP-DM, obteniendo un modelo que permite definir y calificar a un grupo de grandes contribuyentes según su perfil de riesgo. Buscando seguir con esa línea, este trabajo buscará a través de un análisis constructivo cuasi-experimental, focalizar la labor tributaria computacional del Estado esta vez sobre contribuyentes chicos y medianos de acuerdo a la realidad digital y fiscal contemporánea.

## 3.1. LAS FINANZAS PÚBLICAS Y EL CAPITAL SOCIAL COMUNITARIO

La función del Estado, como sujeto de las finanzas públicas para toda comunidad organizada es absolutamente protagónica, tanto que no es concebible la economía de mercado sin la coexistencia de servicios públicos y bienes de capital creados por el Estado para satisfacer las necesidades humanas [Jarach, 1996]; sin embargo tal labor no ha sido fácil, puesto que dicha función ha conllevado a la aparición de complejos problemas de distribución en la práctica así como en el estudio de las leyes que rigen la distribución de los recursos [Ricardo, 1985], ambos han sido siempre un problema primordial de la economía política.

Afortunadamente, abundante literatura iluminó este campo del arte a lo largo de los últimos siglos, la ciencia de las finanzas públicas se ha ocupado de desarrollar diversas teorías para repartir la carga tributaria entre los integrantes de la comunidad, la teoría de la distribución de la carga pública sostiene que la obligación de contribuir es consecuencia de la solidaridad social de todos los miembros de la comunidad, la cual asume el deber de sostenerla. Esa obligación obedece a la capacidad personal del individuo, conforme a sus posibilidades, para sostener los gastos de la comunidad [Garcia Vizcaino, 2009].

A fin de establecer doctrinalmente en qué casos se cumple con el principio de equidad, han sido elaboradas distintas teorías, tales como las del «beneficio», del «sacrificio» y de la «capacidad contributiva», esta última –el de la *capacidad contributiva*– propicia como fundamento de los gravámenes exigidos por el Estado en virtud de su poder de imperio. Dicha tendencia considera la capacidad económica global de los individuos para concurrir al mantenimiento del Estado, exteriorizándose objetivamente a través de manifestaciones ciertas y concretas, verdaderos índices directos o indirectos de riqueza [Garcia Vizcaino, *op. cit.*].

Por otro lado, las finanzas públicas parecen cumplir también una función ética en cuanto a la imposición de los tributos. Esta noción de «ética fiscal» implica por un lado, que la distribución de las cargas deben hacerse equitativamente y por el otro, que el Estado no exija más de lo necesario para cumplir sus funciones; el Estado tiene el deber y la obligación moral de establecer un sistema justo de impuestos, que busque el bien común; por otra parte, los ciudadanos tienen la obligación de pagarlos en tiempo y forma. Por estos motivos, el deber de contribuir a las cargas sociales es una obligación vinculante en conciencia, una noción de ética fiscal fundada en la misma existencia de las comunidades humanas; es el precio mismo de vivir en una comunidad organizada [Celdeiro, 2011] con un capital social embebido que sostiene y estructura dicha organización.

El desarrollo del capital social de una población resulta central en el desarrollo de toda comunidad, existe evidencia de significativas correlaciones entre el grado de confianza existente en una sociedad y eficiencia judicial, ausencia de corrupción, calidad de la burocracia y el cumplimiento de los impuestos [Kliksberg, 2008]; por ende, todo esfuerzo desde las finanzas públicas dirigido a mejorar la eficiencia fiscalizadora del Estado contribuiría efectivamente a acelerar tal efecto multiplicador, dado que todas las acciones destinadas al pago de impuestos representan una manifestación del grado de cohesión social interna de una comunidad, por ende toda acción pública destinada a reducir la desigualdad incrementa el capital social y contribuye al bienestar general [Sen y Klikberg, 2010]. Dos aspectos esenciales deben ser considerados: la conducta del contribuyente y el comportamiento de la actividad fiscal y financiera del Estado, este trabajo intenta abordar herramental informática capaz de analizar dichas conductas a través de la labor del investigador tributario para así enriquecer el capital social comunitario.

## 4. PROBLEMAS E HIPÓTESIS

En Argentina el organismo del Estado encargado de la fiscalización tributaria de sus ciudadanos le corresponde a la Agencia Federal de Ingresos Públicos (AFIP) esta última lleva adelante todas las tareas del sistema de recaudación impositiva así como hace a la fiscalidad entre el Estado y la sociedad. La Administración Federal de Ingresos Públicos tiene la misión de administrar la aplicación, percepción, control y fiscalización de los tributos nacionales, los recursos de la seguridad social y las actividades relacionadas con el comercio exterior; promoviendo el cumplimiento voluntario, el desarrollo económico y la inclusión social [AFIP, 2011]. No obstante, como sucede en toda sociedad organizada, la captación de recursos materiales sobre agentes económicos heterogéneos genera indefectiblemente múltiples formas de oposiciones, resistencias, evasiones o elusiones que socavan la propia razón de ser de la institución.

Estas tensiones de fiscalidad en torno a la función tributaria del Estado se encuentran atravesadas por dimensiones culturales, políticas y económicas estrechamente imbricadas tanto en organizaciones, instituciones como en individuos [Roig, 2008], las cuales muchas veces se expresan en la forma de evasión y/o elusión impositiva lo que presupone un problema para el conjunto de la sociedad toda puesto que socavan su tejido social. Por lo tanto en este sentido la labor del investigador tributario pasa a tener un papel importante en lograr interpretar y ajustar tales tensiones en el entorno social al que este pertenece con impacto en el capital social de su comunidad.

Por otra parte, aunque la organización fiscalizadora del Estado posea un centro de cómputos de avanzada uno de los motores computacionales para la investigación tributaria más potentes con que cuente la Nación [Garello, *op. cit.*], –ver sección 3–, los recursos de dicho complejo se encuentran generalmente enfocados a los grandes contribuyentes así como en tareas operativas masivas relacionadas al plan estratégico de la AFIP, y/o investigaciones complejas de triangulación internacional, dejando de lado la labor de campo descentralizada y artesanal llevada a cabo por el investigador tributario federal situ en las provincias y municipios del interior; estos últimos, por lo general enfocados a las actividades de contribuyentes chicos o medianos, pero igualmente complejos regionalmente, carecen del conocimiento y las capacidades computacionales que les permitan descubrir las reglas y las aglomeraciones del crimen organizado en su situación particular.

Intentando arrojar un poco de luz sobre estas problemáticas identificamos las siguientes hipótesis.

> **[i]:** Es factible el uso de herramientas computacionales para el descubrimiento de actos contra la hacienda pública, patrones de evasión y/o elusión en el contribuyente chico y mediano.
> **[ii]:** Puédase empoderar al investigador tributario no informático con herramientas computacionales que mejoren la eficacia de su labor local y regional.

**[iii]:** Existe una propuesta metodológica, desde la explotación de la información, que permita alguna estrategia heurística viable para el investigador tributario no informático.

**[iv]:** En última instancia, estas técnicas pudieran contribuir a la consolidación del capital social de una sociedad, al mejorar la transparencia y la función fiscalizadora del Estado.

Como es de esperar, no todos los inspectores de la administración federal tienen acceso directo a poder procesar y analizar computacionalmente grandes cantidades de información utilizando el centro de cómputos de la AFIP a mero pedido particular del analista, de allí que, para la labor más regional del investigador tributario federal se propone en estas páginas un enfoque computacionalmente limitado. El desconocimiento de los conocimientos informáticos suficientes para el entendimiento de los procesos computacionales es también otra barrera con la que el investigador del fisco debe lidiar, y que en este esfuerzo se intenta subsanar[5].

# 5. METODOLOGÍA DE EXPLOTACIÓN DE LA INFORMACIÓN

Siguiendo literatura de frontera en el campo del arte de la explotación de la información, se propone el uso de metodología propuesta en Britos y Garcia-Martinez [2008, 2009] para la aplicación de métodos de sistemas inteligentes, para descubrir y enumerar patrones presentes en la información, los cuales cuentan son capaces de obtener resultados de análisis de la masa de información que los métodos convencionales no logran alcanzar [Michalsky, Britos y García Martinez; 1983, *op. cit.*].

La metodología propuesta consta de la combinación de un conjunto de técnicas para procesos de explotación de la información, las cuales se sustenta en la intercalación heurística de distintas algoritmias de sistemas inteligentes: Algoritmos de inducción TDIDT (TDIDT), Mapas auto-organizados de Kohonen (SOM) y Redes Bayesianas o de creencias (RB); los cuales posibilitan –respectivamente– el descubrimiento de reglas de comportamiento, descubrimiento de grupos y el descubrimiento de atributos significativos respectivamente. A *posteriori* también posibilitan el proceso de explotación en combinando la algoritmia en distintos pasos, *e.g.* descubrimiento de reglas de pertenencia a grupos (SOM y TDIDT), ponderación de reglas de comportamiento (TDIDT y RB), ponderación de pertenencia a grupos (SOM y RB), etc. Para ahondar en la temática, se ofrece a continuación una breve descripción de cada algoritmo[6].

## 5.1. ALGORITMOS DE INDUCCIÓN TDIDT

Los árboles de decisión por inducción arriba-abajo TDIDT (Top-Down Induction Decision Trees) son algoritmos de inducción para la toma de decisiones TDIDT son sistemas inteligentes de aprendizaje supervisado capaces de descubrir *reglas de comportamiento* ocultas en los cúmulos de datos e información, son ampliamente usados en el descubrimiento de conocimiento.

El algoritmo TDIDT como parte de la familia de algoritmos de inducción, surgieron a partir de las investigaciones en inteligencia artificial y aprendizaje automático de Quinlan [1986, 1990, 1993], el cual permitió el desarrollo de estos algoritmos descripciones simbólicas de los datos para diferenciar entre distintas clases a partir del cálculo de ganancia de información.

---

[5] En la sección E del Anexo se confeccionaron sendos glosarios especializados para facilitar la tarea del lector ajeno al campo del arte.

[6] Para una explicación más acabada de cada algoritmo repasar los Anexos B, C y D.

En la explotación de la información aplicada existen básicamente dos tipos de algoritmos TDIDT, Steinberg [*op. cit.*] describe sus diferencias como sigue:

> **[ID3]** La elección del algoritmo ID3 apunta a la obtención de un árbol de decisión que explique cada instancia de la secuencia de entrada de la manera más compacta posible o, dicho de otra manera, que de la forma más sencilla separe mejor los ejemplos; el algoritmo, recursivo y sin retropropagación, en cada momento elige el mejor atributo —previa discretización de las variables continuas– dependiendo de una determinada heurística, y favorece indirectamente a aquellos atributos con muchos valores, los cuales no tienen que ser los más útiles.
>
> **[C4.5]** El algoritmo C4.5 genera un árbol de decisión mediante la estrategia de «profundidad primero» (depth-first) a partir de los datos y mediante particiones recursivas. En cada nodo se considera todas las pruebas posibles que pueden dividir el conjunto de datos y selecciona la que resulta en la mayor ganancia de información.

Aunque ambos algoritmos funcionan bajo el concepto de maximización de la ganancia informativa por reducción de entropía la elección del algoritmo C4.5 apunta a la construcción de un árbol no binario y menos frondoso que en el caso del ID3. Gráficamente ambos algoritmos pueden representarse como un tipo particular de grafo acíclico dirigido.

## 5.2. REDES NEURONALES: MAPAS AUTO-ORGANIZADOS DE KOHONEN (SOM)

Lo que empezó como el desarrollo de un nuevo concepto computacional que intentaba modelar la misma estructura neurológica de nodos y aristas en donde la información se distribuye por toda la red, como propuso a partir de Minsky y Papert [1969] en su trabajo seminal en la materia, y continuado por innumerables informáticos, se llegaron a desarrollar sistemas extremadamente eficaces para resolver problemas complejos capaces de manejar enormes cantidades de data. Esto es posible, a partir de la concepción misma de todo sistema neuronal, en el cual, en caso de que un conjunto de nodos no funcione, el sistema todo no se verá afectado en su funcionalidad agregada.

Una red neuronal exige un aprendizaje, de donde se desprende la taxonomía básica de este tipo de inteligencia artificial, esta, como todo sistema inteligente en informática puede ser, **[i]**: una red de aprendizaje supervisado o **[ii]**: una red de aprendizaje no supervisado. En donde en el primer caso, se requiere de la intervención de un operador humano que «entrene» el sistema; o en el segundo caso, donde ninguna interacción humana es requerida para que el sistema pueda aprender, ese lo hace por sí mismo. Este último caso mucho más útil cuando lo que se quiere es trabajar con importantes cantidades de datos.

Los mapas auto-organizados de Kohonen o «SOM» (acrónimo en inglés de Self-Organizing Maps) pertenecen al segundo grupo, un algoritmo no supervisado para el *descubrimiento de grupos*, estas redes permiten el descubrimiento de conocimiento oculto a través de métodos de aprendizaje no supervisado en redes neuronales donde los datos se agrupan a partir de características similares. Estos mapas SOM auto-organizados pueden ser aplicados a la construcción de particiones de grandes masas de información y poseen la ventaja de ser tolerantes al ruido y la capacidad de extender la generalización al momento de necesitar manipular datos nuevos [Kohonen, 1995].

## 5.3. REDES DE CREENCIAS (RB)

Otra tecnología basada en sistemas inteligentes para descubrir conocimiento, son las llamadas redes de creencias o bayesianas (RB), en informática estas redes son utilizadas para calcular la *ponderación de interdependencia entre atributos*. Como otras técnicas de inteligencia artificial, las RB proveen al actor humano de una forma compacta de representación del conocimiento muchas veces oculto en las estructuras de información, lo que resulta en un método flexible de razonamiento [Felgaer *et al.*, 2003]. El obtener una red bayesiana a partir de datos, es un proceso de aprendizaje que se divide en dos etapas: el aprendizaje estructural y el aprendizaje paramétrico.

Básicamente las redes bayesianas son grafos dirigidos acíclicos cuyos nodos representan variables aleatorias en el sentido de Bayes, donde las aristas representan dependencias guardando allí una direccionalidad. Una red bayesiana representa relaciones causales en el dominio del conocimiento a través de una estructura gráfica y las tablas de probabilidad condicional entre los nodos; por lo tanto, el sistema de conocimiento que representa la red está compuesto por los siguientes elementos:

> **[i]:** Un conjunto de nodos $X_i$ que representan cada una de las variables del modelo, cada uno con un conjunto de estados $x_i$.
> **[ii]:** Un conjunto de enlaces o arcos $(X_i; X_j)$ entre aquellos nodos que tienen una relación de causalidad.
> **[iii]:** Una tabla de probabilidad condicional asociada a cada nodo $X_i$.

Donde mientras el postulado **[i]** asegura un conjunto de estados mutuamente excluyentes, **[ii]** manifiesta que todas las relaciones del sistema estén explícitamente representadas por el grafo, y el **[iii]**, que indica la probabilidad del estado de cada nodo $X_i$ condicional a cada combinación de estados de los nodos ancestros a este, a menos que $X_i$ solo tenga nodos sucesores en ese caso la probabilidad es *a priori*.

A pesar de su nombre, las redes bayesianas o de creencias no implican necesariamente el uso de estadística Bayesiana, sin embargo es llamado así por la utilización de la regla de Bayes para el cálculo de inferencia probabilística –llamarlo como un gráfico acíclico dirigido sería quizás más apropiado– dado que un clasificador de Bayes ingenuo (Naive Bayes) resulta útil para la representación de modelos jerárquicos condicionados en algún atributo.

# 6. APROXIMACIÓN HEURÍSTICA A UN CASO HIPOTÉTICO

La imposibilidad de acceder a información de contribuyentes reales –por restricciones legales de acceso a datos personales– que posibilite aproximación a un caso de uso así como la necesidad de ejemplificar la aplicación de algoritmos obligan a encarar una aproximación pragmática para la correcta validación del problema de investigación fiscal.

Considerando la imposibilidad de acceder a la información tributaria real a la que apunta este trabajo, dado que esta es solo accesible a los investigadores tributarios de la administración federal siendo estos penosamente castigados en caso de exponer información privada[7], se decidió menester propiciar una aproximación hipotética, mediante una base de datos no real pero análoga a aquella que pueda resultar familiar a todo investigador tributario con oficio en la detección de evasores fiscales.

---

[7] El artículo No. 157 del Código Penal Argentino establece penas de prisión desde un mes a dos annos a todo funcionario público que revelare hechos, actuaciones, documentos o datos, que por ley deben ser secretos.

La aproximación a una base de datos hipotética con la que pueda contar un investigador tributario de la AFIP se confeccionó a partir de los datos con los que suele contar la institución tales como el padrón de contribuyentes, presentación de declaraciones juradas, sistema de seguimiento de juicios, sistema de seguimiento de fiscalizaciones, el Banco Central de la República Argentina, imprentas autorizadas a emitir facturas, registros, etc.

La aproximación empírica hipotética toma la nomenclatura proveniente de ingeniería de base de datos [Bertone y Thomas, 2012], donde se asume que el investigador tributario posee en su disco el archivo plano de una base de datos de dimensiones modestas, denominada «contribuyentes», representativa esta de un subconjunto del universo de la totalidad de contribuyentes en su respectiva área de responsabilidad fiscal, la cual puede expresarse de la siguiente manera[8]:

> contribuyentes =    (#CUIT/CUIL, razonsocial, perfil,
> tipopersona, apellido, apellidomater,
> nombres, fechanac, estado, rellab-
> condiciónIVA, categmonot, decjurada,
> liquidez, asinpresdj, superpodom,
> supdompjur, supdompfis, blanque-morat,
> accionista, accmayorit, directivosoc,
> donayoacred, nroemple, nrodenuncias,
> cantcau, contribsan, siper)

Esta tabla –o *relación* en la jerga del arte– está conformada por datos de personas experimentales no reales de un conjunto de uso acotado a 114 registros y 24 atributos hipotéticos, partimos de registros de características normales con los que cuenta y tendría exceso un inspector tributario.

Estas observaciones –*registros* en la jerga informática– suelen encontrarse plagadas de valores nulos y/o desconocidos lo que plantea un problema para el experto informático tributario en la preparación previa de la data, por lo que las bases deben previamente puestas a punto creando nuevos atributos que posibiliten la interpretación de los mismos. Al momento de usar la herramienta computacional *Tanagra*[9] es importante hacer una buena preparación de los datos puesto que la tabla –o *relación* en teoría de base de datos– no debe contener celdas vacías y muchas veces hay reducir la cantidad de atributos.

---

[8] Para una descripción de cada variable tributaria –*atributos* en la jerga informática– ver el glosario de atributos del Anexo E.
[9] En este caso fue usada la versión *open source* Tanagra 1.4.25. También son recomendables otros paquetes de código abierto para minería de datos tales como *RapidMiner* o *Weka*.

## 6.1. ESTRATEGIA HEURÍSTICA

A partir de la herramienta computacional de minería de datos se intentan plasmar distintas combinaciones algorítmicas de acuerdo a Britos y Garcia-Martinez [2009] el explotador de la información puede encontrarse con las siguientes estrategias a partir de su propia heurística investigativa.

$$\begin{aligned} &\text{SOM} \xrightarrow{priori} \text{RB} \\ &\text{TDIDT} \xrightarrow{priori} \text{RB} \\ &\text{TDIDT} \xrightarrow{priori} \text{SOM} \xrightarrow{posteriori} \text{RB} \\ &\text{SOM} \xrightarrow{priori} \text{TDIDT} \xrightarrow{posteriori} \text{RB} \\ &\text{TDIDT} \xrightarrow{priori} \text{RB} \xrightarrow{posteriori} \text{SOM} \xrightarrow{fortiori} \text{RB} \\ &\text{SOM} \xrightarrow{priori} \text{RB} \xrightarrow{posteriori} \text{TDIDT} \xrightarrow{fortiori} \text{RB} \end{aligned} \quad (1)$$

De acuerdo a la base de datos se adoptó la siguiente estrategia heurística de tres pasos de acuerdo a las opciones metodológicas disponibles.

$$\text{SOM} \xrightarrow{priori} \text{TDIDT} \xrightarrow{posteriori} \text{RB}$$

Estrategia de abordaje que supone el procesamiento de datos con sistemas inteligentes a través de tres pasos. Un descubrimiento de reglas de pertenencia a grupos *a priori* y *posteriori*, –usando redes neuronales (SOM) y algoritmos TDIDT– con ponderación de reglas de comportamiento *a fortiori*.

La red neuronal SOM adoptada tomó la siguiente configuración:

SOM = (fechanac, asinpresdj, nroemple, nrodenuncias, cantcau)

De donde es necesario recordar que esta solo toma valores continuos, lo que representa una restricción importante del algoritmo. Del conjunto de datos resultó dividido en cuatro agrupamientos de *clusters* como se desprende de la siguiente tabla.

Tabla 1: **Topología del mapeo SOM**

|  | $CSOM_{i1}$ | $CSOM_{i2}$ |
|---|---|---|
| $CSOM_{1j}$ | $CSOM_{11}(57)$ | $CSOM_{12}(16)$ |
| $CSOM_{2j}$ | $CSOM_{21}(32)$ | $CSOM_{22}(9)$ |

Donde los grupos $CSOM_{11}$ y $CSOM_{21}$ resultaron los más numerosos del mapeo. Ahora es necesario descubrir cuáles son las reglas ocultas que hacen a esos grupos, por lo que el algoritmo de inducción del siguiente paso debe estar dirigido al nuevo atributo «CSOM» generado por la red neuronal como atributo clase u objetivo, tal como se desprende de la siguiente configuración de algoritmos.

TDIDT (C4.5) = (**CSOM**, perfil, tipopersona fechanac,
estado, rellab-condiciónIVA, categmonot,
decjurada, liquidez, asinpresdj,
superpodom, supdompjur, supdompfis,
blanque-morat, accionista, accmayorit,
directivosoc, donayoacred, nroemple,
nrodenuncias, cantcau)

En consonancia con lo enunciado por Steinberg, para el caso de aplicación de árboles de inducción para el descubrimiento de reglas de comportamiento, observamos que una configuración con algoritmo C4.5 resultó más eficaz dado que se pudo obtener más generación de información con mayor frondosidad.

El algoritmo TDIDT C4.5 fue configurado para descubrir las reglas de pertenencia a los agrupamientos descubiertos en el primer paso. De esta manera fue posible hallar las siguientes reglas como se desprende de la Tabla 2.

Tabla 2: **Ganancia de información (GI)**

| $GI(A,S)_0$ | $GI(A,S)_1$ | $GI(A,S)_2$ | $GI(A,S)_3$ |
|---|---|---|---|
| ○: contrib | | | |
| | •: fechanac <1972.5 | | |
| | | ○: nroemple <16 | |
| | | | ■: asinpresdj <2.0 ∈ $CSOM_{21}$ 100 % de 11 obs. |
| | | | ■: asinpresdj ≥ 2.0 ∈ $CSOM_{12}$ 83 % de 6 obs. |
| | | ○: nroemple ≥ 16 ∈ $CSOM_{22}$ 100 % de 6 obs. | |
| | •: fechanac ≥ 1972.5 | | |
| | | ○: supdompjur [NO] ∈ $CSOM_{11}$ 95 % de 60 obs. | |
| | | ○: supdompjur [SI] ∈ $CSOM_{12}$ 100 % de 11 obs. | |

De la algoritmia presentada *ex supra*, se induce que haber nacido antes o después de 1972 –«fechanac»– podría tener alguna relevancia como regla de comportamiento en la pertenencia a alguno de los agrupamientos descubiertos; puesto que para aquellos nacidos antes de 1972, poseer 16 o más empleados –«nroemple»– a cargo del contribuyente parece condicionar la pertenencia al grupo $CSOM_{22}$, mientras que la pertenencia a los grupos $CSOM_{21}$ y $CSOM_{12}$ parece estar condicionada a contribuyentes con menos de 16 empleados que –en una tercera partición de la información– con menos de 2 o más de 2 períodos anuales sin haber presentado declaración jurada alguna respectivamente –«asinpresdj»–. Por otro lado, para los nacidos posteriormente a 1972 la existencia de superposición de domicilios de contribuyentes con personas jurídicas parece significar una regla de pertenencia a los grupos $CSOM_{11}$ y $CSOM_{12}$ –«supdompjur»–.

Gráficamente la ganancia de información del algoritmo C4.5 puede plasmarse en el siguiente grafo de la Figura 2.

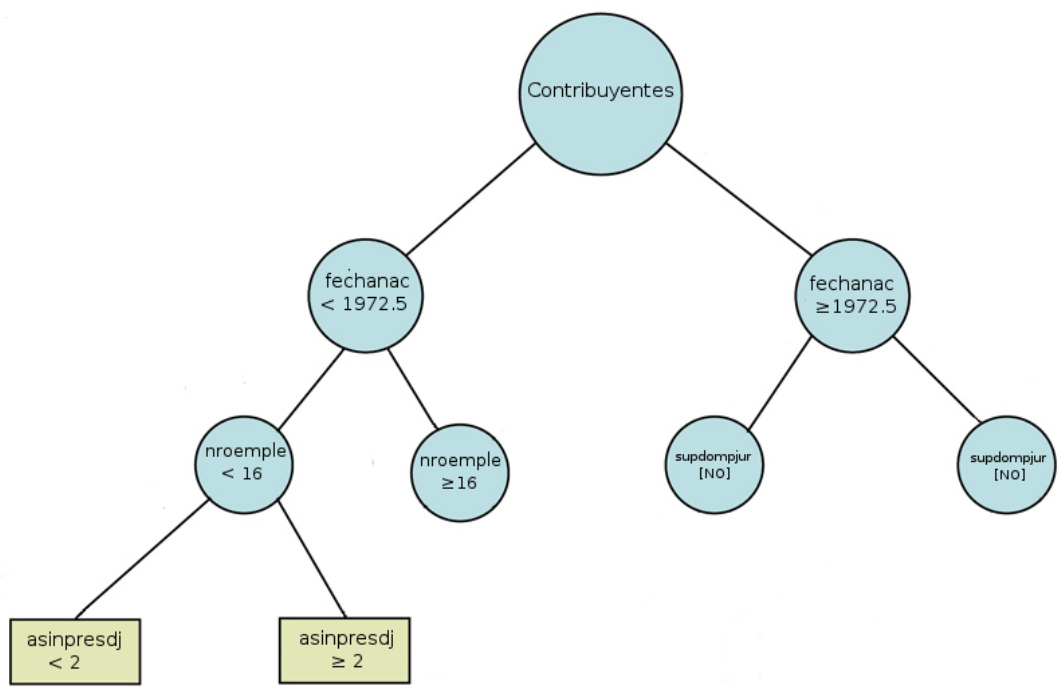

Fig. 2: Árbol de inducción por algoritmo TDIDT (C4.5)

A *fortiori*, como último paso de la estrategia heurística adoptada, se aplica la siguiente configuración de una red de creencias mediante un *naive Bayes* condicional en el atributo «supdompjur», *i. e.* existencia de contribuyentes con superposición de domicilios con personas jurídicas[10].

> RB = (**supdompjur**, estado, rellab-condiciónIVA,
> categmonot, decjurada, liquidez,
> superpodom, supdompjur, supdompfis,
> blanque-morat, accionista, accmayorit,
> directivosoc, donayoacred)

Intuitivamente, se decidió trabajar sobre la probable causalidad que podrían tener los atributos «liquidez», «blanque-morat», «accionista», «donayoacred» sobre la existencia de superposición domiciliaria con personas jurídicas; los que una vez corrido el algoritmo clasificador bayesiano ingenuo, se obtuvieron los siguientes resultados de probabilidad condicional.

---

[10] En el caso del atributo «supdompjur», en otros casos de aplicación, podría propiciarse el uso de algoritmos de inducción borrosos, puesto que en muchos casos la persona jurídica o física sospechada declara un domicilio espacialmente cercano al de otra persona presumida de perjurio fiscal.

Tabla 3: **Probabilidad condicional: superposición domicilio persona jurídica**

| atributo supdompjur | liquidez | | | blanque-morat $P(X_2)$ | accionista $P(X_3)$ | donayoacred $P(X_4)$ |
|---|---|---|---|---|---|---|
| | $P(X_{1,Baja})$ | $P(X_{1,Media})$ | $P(X_{1,Alta})$ | | | |
| P(NO\|X) | 0.021 | 0.887 | 0.083 | 0.299 | 0.052 | 0.010 |
| P(SI\|X) | 0.941 | 0.000 | 0.059 | 0.000 | 0.941 | 1.000 |

Como se desprende de la Tabla 3 ilustra que la existencia de superposición de domicilio parece estar condicionada a una baja liquidez relativa entre el acervo patrimonial declarado y la disponibilidad líquida del contribuyente, no haber entrado en moratorias o blanqueos de capitales, poseer acciones y haber otorgado créditos y/o haber hecho donaciones. Ahora en la forma de un gráfico acíclico dirigido (GAD) se plasman los resultados en la fase gráfica, interpretando una estructura de la red creencias de la siguiente forma.

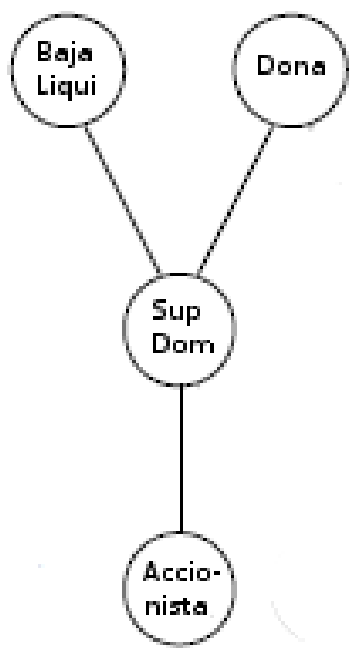

Fig. 3: GAD de red de creencias

La estructura de la red Bayesiana, que se desprende de la Figura 3, presenta el soporte gráfico de la red de creencias basadas en la Tabla 3 pivotada en el atributo «supdompjur», que en la net se entiende intuitivamente condicional a la existencia de contribuyentes con baja liquidez relativa y haber efectuado donaciones. A posteriori, se intuye la existencia de contribuyentes con superposición de domicilios con personas jurídicas como evento causal a la posesión de acciones en el acervo patrimonial de los mismos.

# 7. CONCLUSIONES

En este trabajo se describió una breve introducción a conceptos básicos de inteligencia artificial para el entendimiento de los procesos de explotación de la información, se exploraron técnicas de algorítmica computacional de frontera aplicados a un caso hipotético de aplicación tributaria fiscal como caso de validación, a partir de la elaboración de un archivo plano trivial confeccionado desde la óptica y el oficio del inspector fiscal con capacidades computacionales limitadas.

Por otra parte, el trabajo intento contribuir con un caso de uso de tecnologías de sistemas inteligentes en la administración tributaria ampliando a partir de lo alcanzado por Steinberg [*op. cit.*] aunque facilitando el uso de técnicas de explotación de la información para la búsqueda de patrones de delito fiscal en contribuyentes de dimensiones chicas y medianas, así como también, abrir una puerta al investigador fiscal no informático hacia nuevos herramentales. El planteamiento del caso de validación hipotético estuvo sujeto a las posibilidades fácticas de la herramienta la que se utilizó en base a una propuesta metodológica de técnicas en procesos de explotación de información.

En el proceso se partió de una estrategia de explotación de la información con descubrimientos de grupos *a priori*, reglas de pertenencia a estos grupos *a posteriori* y ponderación de reglas de comportamiento de pertenencia de atributos *a fortiori*, se corroboraron algunos de los postulados arribados por Steinberg en cuanto al uso de los algoritmo de inducción C4.5 se vuelve una opción factible de elección a la hora de buscar reglas de comportamientos ocultas en los datos.

Finalmente, como es habitual en la literatura informática computacional, las futuras líneas de investigación resultan una alusión de notable relevancia, dada la vertiginosidad e interdisciplinariedad del avance es esta área del conocimiento: en economía y finanzas públicas, la utilidad de las técnicas provistas al investigador tributario podrían ir más allá de la utilización de redes neuronales, redes bayesianas y/o algoritmos de inducción; por ejemplo, si se contase con el relevamiento adecuado de los procesos en los que interviene el investigador tributario, podría implementarse sistemas de expertos[11] y/o técnicas de aprendizaje automático. Todas alternativas dignas de ser exploradas con ahínco, siendo que pudieran resultar en la consolidación del capital social de una nación.

---

[11] Forma de inteligencia artificial que emula al experto humano dentro de un campo acotado, lo cual agiliza la toma de decisiones suplantando la experiencia del experto en sistemas multiagentes.

# REFERENCIAS

# ANEXO

## A. ASPECTOS BÁSICOS SOBRE TEORÍA DE GRAFOS

Dado que los algoritmos utilizados en inteligencia artificial, ya sean redes neuronales, redes de creencias y árboles de inducción son representados analíticamente por medio de grafos, es necesario suplir al lector no especializado con una breve introducción sobre el tema.

**Definición 1** Un grafo $G$ es un par de conjuntos $G = (X; A)$ t.q. $X$ es un conjunto de $n$ nodos donde $n \in X$ y $A$ con un conjunto de $m$ arcos o aristas t.q. $|X| = n$; $|A| = m$.

**Definición 2** Un nodo $X$ es una variable aleatoria que puede tener varios estados $x_i$. La probabilidad de que el nodo $X$ este en el estado $x$ se manifiesta como $P(x) = P(X = x)$.

**Definición 3** Una arista o arco, se define por un par ordenados de nodos $(X; Y)$ por lo que representa la dependencia entre dos variables del sistema.

**Definición 4** El nodo $X$ es ancestro del nodo $Y$ si existe un arco $(X; Y)$ entre los ambos.

**Definición 5** El nodo $Y$ es descendiente o sucesor del nodo $Y$ si existe un arco $(X; Y)$ entre los ambos.

## B. ÁRBOLES DE INDUCCIÓN TDIDT

Los algoritmos Top Down Induction Decision Tree (TDIDT) -árboles de decisión por inducción arriba-abajo- permiten el desarrollo de descripciones simbólicas de los datos para diferenciarlos entre distintas clases [Quinlan, 1990] lo que permite al humano que entrena el algoritmo descubrir reglas de comportamiento ocultas en los datos. Estos pertenecen a los métodos inductivos de aprendizaje automático que aprenden por si mismos a partir de ejemplos preclasificados.

Entre los algoritmos de inducción para la toma de decisiones más utilizados se encuentran los algoritmos ID3 y C4.5 –el segundo como superador del primero– ambos desarrollados por Quinlan [1986, 1993]. Estos recurren al cálculo de la ganancia de información que genera la disminución entrópica en conjunto dado de información al particionarse la data en función de una determinada característica o atributo.

Esta ganancia de información ($GI$) se estima a través del cálculo de la entropía ($E$) – asunción provista por la teoría de la información– la cual intenta cuantificar la reducción esperada de entropía al particionar las observaciones de acuerdo a un atributo determinado, donde la ganancia de información será función de un atributo «$A$» relativo a la colección de información y las instancias de dicho atributo «$S$», guardando la siguiente relación en la ecuación (2).

$$GI(A, S) = E(S) - \underbrace{\sum_{v \in A} \frac{|S_v|}{|S|} E(S_v)}_{I(M)} \qquad (2)$$

Donde la entropía, representada por $E(S)$, caracteriza la pérdida o impureza de una colección arbitraria de información, $v$ un valor del atributo o variable $A$ del conjunto de datos, $S$ el número de instancias o categorías dentro de cada atributo y $S_v$, el subconjunto de instancias en donde $A$ toma el valor $v$.

Por teoría de la información se tiene entonces que $I(M)$ se entiende como el contenido total de información, tal que un universo de mensajes denotado como $M = (m_1, m_2, ..., m_n)$ donde cada mensaje $m$ tiene una probabilidad $p(m_i)$ de conformar el conjunto informativo, sucintamente el contenido de la información se define como sigue:

$$I(M) = -\sum_{i=1}^{n} -p(m_i) log_2(p(m_i)) \qquad (3)$$

El contenido de la información de la ecuación (3) se expresa en números de bits, la cantidad de información generada es equivalente a la entropía perdida por el sistema en sentido inverso, por ende, cuanto mayor información sea obtenida por la partición menor entropía del subconjunto obtenido.

De esta forma a partir del cálculo de la ganancia de información para cada atributo, se seleccionará el atributo que otorgue, según la ecuación (2), la mayor ganancia absoluta en bits y/o la menor generación de entropía correspondiente a tal ganancia. De esta forma, a partir de aquel atributo seleccionado desde un nodo madre o raíz del árbol de inducción –*q. e.* desde el conjunto total de información sin particionar– inducirá la partición de la data de acuerdo a la cantidad de instancias de valores que dicho atributo posea. Lo que, al correr el algoritmo recursivamente, a través de los subconjuntos que se van creando, se obtiene para cada nodo una partición con una entropía cada vez más cercana a cero. En la Figura 4 puede apreciarse este proceso de ganancia de información y reducción de entropía.

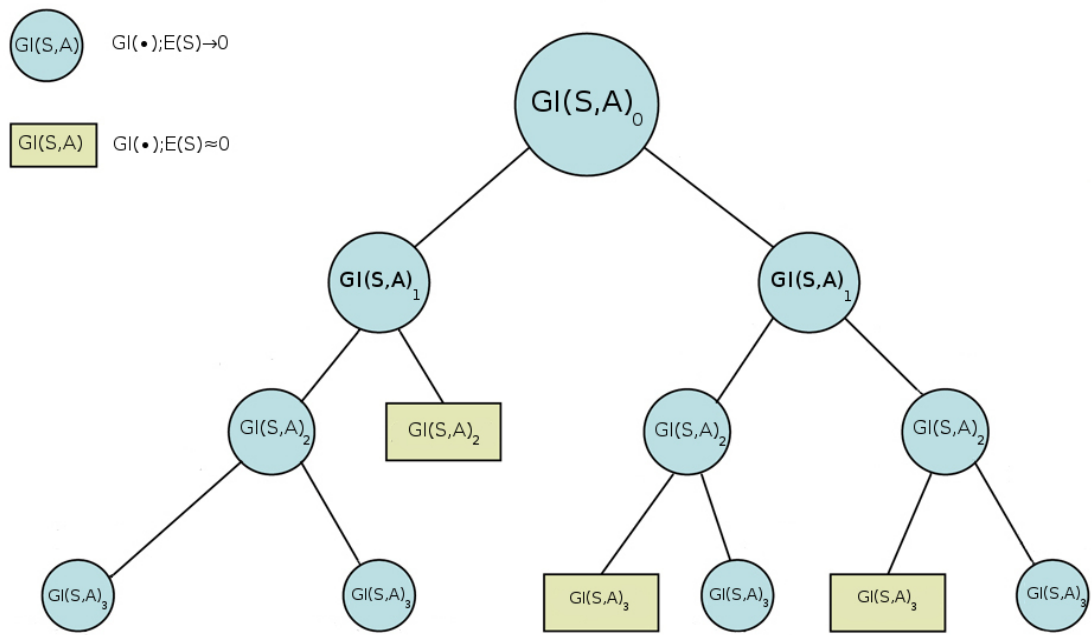

Fig. 4: Árbol de inducción por algoritmo de ganancia entrópica de información

Donde se aprecia claramente la ganancia de información progresiva que se obtiene en el sistema a medida que la entropía tiende a cero. Por otra parte, un árbol de inducción también puede ser entendido como un tipo particular de grafo acíclico dirigido.

## C. REDES NEURONALES SOM

El algoritmo de mapeo auto-organizado (SOM – acrónimo en inglés de Self Organizing Maps) propiamente dicho es un algoritmo de inteligencia artificial que trabaja como red neuronal. Desarrolladas por el académico Finlandés Teuvo Kohonen en 1982 [Kohonen, 1995] este tipo particular de red neuronal se caracteriza por mapear por sí mismo están compuesto por un algoritmo de sistema automático de aprendizaje no supervisado, –i.e. que no necesita de un experto humano que entrene la red– de manera que la red neuronal intenta representar en ese mapeo las características de los datos de entrada a la red tal que se preserven las relaciones topológicas presentes en la misma.

A diferencia de otros sistemas inteligentes representados por grafos, en las redes SOM cada nodo del mapa de encuentra enlazado al resto de los nodos, lo que le posibilita representar data multidimensional en pocas dimensiones facilitando la comprensión del analista.

### C.1. EL ALGORITMO SOM

Para conocer el funcionamiento intrínseco del algoritmo de Kohonen, para la detección de cúmulos de datos automáticamente, se encuentra compuesto por fases las cuales pueden descomponerse en seis pasos de la siguiente manera:

**[i]:** Cada nodo ponderado se encuentra inicializado[12].

---

[12] Lo que en informática se entiende como algo en estado de encendido o activado.

**[ii]:** Un vector es elegido aleatoriamente desde un conjunto dado de datos de entrenamiento y presentado.

**[iii]:** Cada nodo en la red es examinado para calcular cuales ponderadores son más probables para cada input. El nodo objetivo o ganador es conocido como la Unidad de Mejor Combinación o *Best Matching Unit* (BMU) –ver ecuación (9)–.

**[iv]:** El radio del entorno del BMU se calcula, *maiore a minus*, a partir de un amplio radio inicial, el radio luego disminuye paulatinamente con cada iteración temporal –ver ecuaciones (5) y (6)–.

**[v]:** Todo nodo dentro del radio del BMU, estimado en **[iv]**, es ajustado para hacer más cercano el mismo al input del vector –ver ecuación (7)–; por lo tanto, cuanto más cerca se encuentre el nodo al BMU, más ponderadores serán alterados –ver ecuación (8)–.

**[vi]:** Se repite sucesivamente a partir del paso **[ii]** para *N* iteraciones.

Las consecuencias automáticas de este algoritmo se fundamentan ecuacionalmente de manera que la ecuación para el cálculo de la distancia entre la data de entrada y los nodos ($DI$) quedan representados de la siguiente forma en la ecuación (4).

$$DI^2 = \sum_{i=0}^{i=n}(I_i - W_i)^2 \tag{4}$$

Donde $I$ representa un vector de inputs, $W$ el vector de ponderadores de un nodo y $n$ el número de ponderación, matemáticamente (4) no es más que la formula de distancia euclidiana aquí en relación a los datos de entrada, distancia elevada al cuadrado lo que simplifica el cómputo y hace a la necesidad de ordenar los valores en una escala uniforme que posibilite la comparación entre cada inputs y su peso

Otra ecuación necesaria para el auto-aprendizaje del algoritmo es la ecuación del radio del entorno tal que:

$$\sigma(t) = \sigma_0 e^{-t/\alpha} \tag{5}$$

Donde $t$ representa la iteración actual, $\sigma_0$ representa el radio del mapeo a partir de un dato de entrada cualquiera y $\lambda$ como la constante temporal de acuerdo a la ecuación (6) descripta *ex infra*.

$$\lambda = \frac{NI}{RM} \tag{6}$$

Siendo $NI$ el número de iteraciones y $RM$ el radio del mapeo sobre la data considerada, el cual es casi arbitrario.

$$W(t+1) = W(t) + \Theta(t)L(t)(I(t) - W(t)) \tag{7}$$

Cual denota la función de aprendizaje principal, siendo $L(t)$ denota el aprendizaje automática del sistema neuronal, donde cada $W(t + 1)$ representa el nuevo valor aprendido de un nodo, que tendrá un mayor peso en tanto el nodo se encuentre más alejado del vector de entrada mayor será el aprendizaje del mismo. Su velocidad, la tasa de tal aprendizaje se presenta a continuación en la ecuación (8).

$$L(t) = L_0 e^{(-t/\lambda)} \tag{8}$$

Donde (8) define cómo aprende cada nodo de manera que cada nodo más cercano al BMU aprende más de los nodos alejados –siempre dentro del radio actual considerado– desde su valor máximo de $\theta = 1$ correspondiente al BMU en sí mismo. Es importante notar que tanto (8) como (5) configuran un declive exponencial puesto que la tasa de aprendizaje automático del sistema decrece a una tasa proporcional a su valor actual a partir de $t = 0$ tanto como se incrementen las iteraciones.

$$\Theta(t) = e^{-DBMU^2/(2\sigma^2(t))} \tag{9}$$

Siendo *DBMU* representa el número actual de nodos entre el nodo actual y la unidad de mejor combinación (BMU), $\theta(t)$ hace que los nodos más cercanos a BMU aprendan más que aquellos nodos en las afueras del radio de mapeo, de donde se desprende la siguiente proposición.

> **Proposición 1** Tanto como se incremente el aprendizaje de un conjunto de nodos $\theta(t) \to 1$ luego $DBMU \to 0$.

Lo que gráficamente puede ilustrarse como sigue.

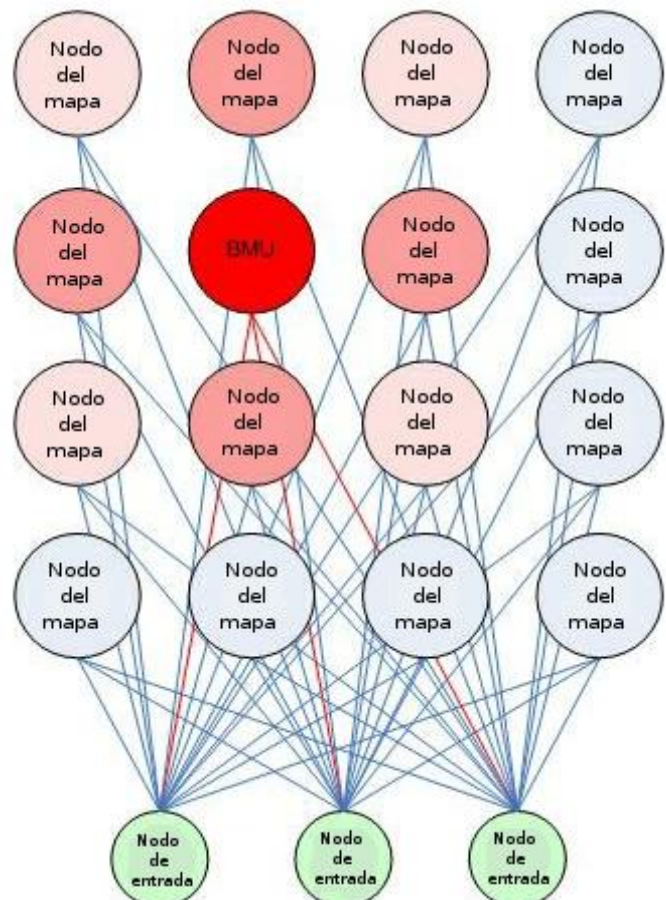

Fig. 5: Red neuronal por mapeo auto-organizado de Kohonen

El **BMU** es seleccionado dentro de una red 4×4 con 4×4×3 = 48 enlaces, donde el paso **[iv]** del algoritmo calcula el radio del entorno, el paso **[v]** aplica la función de aprendizaje a todos los nodos, donde el nodo en el BMU es el que más aprende.

## D. GRAFOS ACÍCLICOS DIRIGIDOS: REDES BAYESIANAS

Una red bayesiana o de creencias está basado en un clasificador Bayesiano ingenuo (Naive Bayes), un simple clasificador probabilístico basado en el teorema de Bayes con supuestos fuertes sobre la independencia condicional de las variables o nodos del sistema. En informática e inteligencia artificial es un proceso de aprendizaje supervisado que se divide en dos etapas: el aprendizaje estructural y el aprendizaje paramétrico [Pearl, 1988]. La primera, consiste en obtener la estructura de la red bayesiana, –*i. e.* las relaciones de dependencia e independencia entre las variables involucradas–, tanto que la segunda tiene como finalidad obtener las probabilidades a priori y a posteriori requeridas a partir de una estructura dada [Felgaer et al., 2003].

La topología –o estructura de una red de creencias– representa tanto las dependencias probabilísticas por un lado entre variables así como describe las independencias condicionales entre las mismas Los nodos representan variables aleatorias que pueden ser continuas o discretas. La estructura de una red bayesiana se construye de acuerdo a las siguientes reglas:

**[i]:** Se asigna un nodo a cada variable $X_i$ indicando de que otros nodos el estado de $X_i$ es causa directa t.q. el conjunto de nodos que causan $X_i$ se lo denomina como conjunto $\pi_X$.

**[ii]:** Se enlazan los nodos con direccionalidad desde sus nodos ancestros (causales) hacia sus nodos descendientes (consecuentes).

**[iii]:** A cada nodo $X_i$ se le asigna una matriz $P(x_i|\pi_{X_i})$ la cual estima la probabilidad condicional de un evento $X_i = x_i$ dada una combinación de ancestros $\pi_X$.

**Proposición 2** Una variable $X_i$ es condicionalmente independiente de una $Y$ si, existiendo una variable $Z$ t.q. $Y \perp\!\!\!\perp X$ –*i.e* $X$ e $Y$ son independientes–, por lo tanto $P(X|Y,Z) = P(X|Z)$.

Lo que facilita la estimación de la probabilidad a posteriori a través del cálculo de la probabilidad conjunta de todas las variables. Puesto que la independencia condicional permite obtener la probabilidad conjunta a partir de las probabilidades condicionales de cada nodo en función de sus ancestros a la vez que simplifican la representación del conocimiento en menos cantidad de parámetros así como en el proceso de inferencia y propagación del conocimiento.

En *verbi gratia* a la proposición 2 se tiene el siguiente ejemplo.

**Ejemplo 1** Un nodo $E$ es independientemente condicional de $A; B; C; D; F; G$ dado $B$ t.q. $P(E|A; B; C; D; F; G) = P(E|B)$.

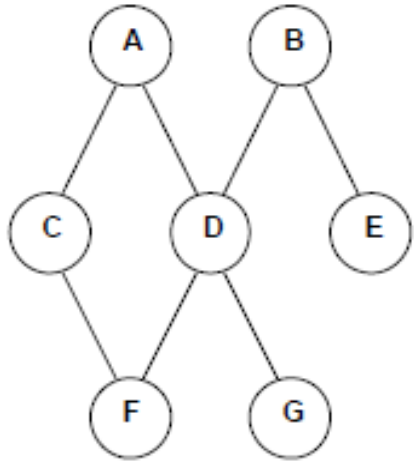

Fig. 6: GAD - Ejemplo de Grafo 1

Donde es fácilmente apreciable la independencia condicional del nodo $E$ respecto de los nodos $A; C; D; F; G$ no así del nodo $B$, el cual representa su ancestro directo.

Si en cambio quisiéramos calcular la probabilidad conjunta de las variables de un sistema siendo conocida la estructura del grafo así como las probabilidades condicionales de cada nodo –*q. e.* variable o registro– en función de sus ancestros, para muchos nodos puede usarse la regla de la cadena para el cálculo conjunto de probabilidad siguiendo a [Pearl, 1988], tal que:

$$P(x_1, x_2, \ldots, x_n) = \prod_{t=1}^{n} P(x_t | x_1, x_2 \ldots, x_{t-1}) = \prod_{t=1}^{n} P(x_t | \pi_{X_t}) \tag{10}$$

Ejemplificando para un caso de uso plasmado en Pearl [*op. cit.*] –una red de creencias que intenta representar la sistémica de la propagación del mensaje que dispara una alarma–. El caso describe a un individuo que percibe el incidente del disparo del sonido de una alarma «$s$» consecuente de dos eventos posibles: un ladrón «$h$» o un terremoto «$e$» –variables explicativas–, asumiendo independencia entre los dos últimos eventos puesto que un anuncio radiofónico «$r$» se confirmaría en la brevedad si hubiese sido el caso de un terremoto. La explicación del incidente se refleja a posteriori del disparo en , a través de la llamada aclaratoria del individuo «$w$», el testimonio del individuo «$g$» o la duda del individuo «$d$» en llamar o no para confirmar la causa del evento.

La estructura del problema, representada en una red creencias plasmada en un grafo $G(X; A)$, contiene los nodos $|X| = (d; e; g; h; r; s; w)$, los cuales deben ordenarse de mayor a menor de acuerdo a la sucesión descendiente expresa en el problema de Pearl; ergo, se determina el siguiente orden tal que $|X| = (h; e; r; s; d; w; g)$ tal y como se aprecia en la estructura del sistema visualizado *ex infra*.

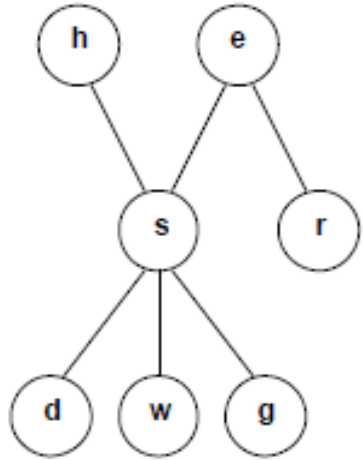

Fig. 7: GAD - Ejemplo de grafo 2

Por lo que conociendo la estructura del grafo puédase calcular la probabilidad condicional conjunta a partir de la ecuación (10) del grafo de la Figura 7 se tendrá la siguiente configuración en (11).

$$P(h,e,r,s,d,w,g) = P(h)P(e)P(r|e)P(s|e,h)P(d|s)P(w|s)P(g|s) \qquad (11)$$

La cual calcula la probabilidad conjunta de todos los nodos del sistema representado por el grafo de la Figura 7, a partir de las probabilidades condicionales de cada uno de los nodos en función de sus ancestros.

## E. GLOSARIOS

### Glosario informático computacional

**algoritmo computacional**: Es la secuencia de instrucciones escritas en algún lenguaje de programación ejecutadas en una computadora, los cuales representan un modelo de solución para un determinado tipo de problema.

**atributo**: En ciencias de la computación, se denomina atributo a una especificación o característica que define la propiedad de un objeto, elemento o archivo en cuestión; bastante lo cual, en términos generales visto como una característica de una propiedad, lo que en ciencias económicas podríamos entender en general como una «variable».

**atributo clase**: En explotación de la información, se denomina atributo clase al seleccionado como variable objetivo sobre el cual aplicar algún algoritmo de inteligencia artificial.

**base de datos**: Una base de datos es un conjunto de datos pertenecientes a un mismo contexto y almacenados sistemáticamente para su posterior uso.

**conocimiento**: En informática, se entiende al conocimiento como el estadio superior de en el proceso de desarrollo de la información.

**CRISP-DM**: Acrónimo en Inglés de Cross Idustry Standard Process for Data Mining, una metodología jerárquica de seis fases que cubre el ciclo de vida para todo proyecto de minería de datos.

| | |
|---:|:---|
| **datos**: | Un dato hace referencia a la unidad mínima de información, los cuales individualmente no son capaces de influir en la toma de decisiones humanas. |
| **explotación de la información**: | Subdisciplina informática que aporta las técnicas y procesos para la transformación de información en conocimiento. |
| **heurística**: | En informática –particularmente en inteligencia artificial– se refiere a los procesos y metodologías no formalizados así como al arte práctico para la resolución de problemas, *q. e.* la búsqueda de los algoritmos efectivos para la resolución de un problema vinculado a un sistema inteligente. |
| **minería de datos**: | Conjunto de procesos y algoritmia usadas para encontrar patrones de información en grandes masas de datos. |
| **nodo**: | En informática, suele referirse a una variable concebida a partir de un registro relacionada a su vez a otros nodos y registros. Puede representarse mediante teoría de grafos, donde un nodo, vértice o punto simboliza una intersección o unión entre dos o más entidades. |
| **red**: | Se denomina así a toda estructura que cuente con un patrón característico. En el campo, una red suele representar estructuras de datos y otras entidades. La teoría de grafos resulta idónea para la representación de tales estructuras. |
| **registro**: | En informática, análogamente a lo que se entiende generalmente como una observación en la jerga del científico no informático, un registro, tupla o fila, representa un objeto único de datos implícitamente estructurados en una tabla o relación. |
| **sistemas inteligentes**: | Se refiere a un programa computacional que reúne características y comportamientos asimilables al de la inteligencia humana o animal. Por ejemplo, algoritmos genéticos, algoritmos TDIDT, redes neuronales BP, redes neuronales SOM, redes Bayesianas, son sistemas inteligentes empleados en la explotación de la información. |
| **tupla**: | En ciencias de la computación una tupla es un objeto[11] que puede tener datos de distintos objetos, –*i.e.* un conjunto de elementos de distintos tipos que se guardan de forma consecutiva en la memoria–. En teoría de base de datos una tupla se define como una función finita que mapea. En teoría de grafos, ampliamente usado en inteligencia artificial, las tuplas se emplean para describir objetos matemáticos que poseen cierta estructura por lo que permiten realizar descomposiciones en subcomponentes. |
| **data warehouse**: | En informática se refiere a una colección de datos orientada hacia algún tipo de organización, contienen la totalidad de los datos de cada sistema operativo de la institución de forma integrada, los cuales a través de *data marts* –subconjuntos específicos del *data warehouse*- sirven para la toma de decisiones en todos los niveles. |

## Glosario tributario

| | |
|---:|:---|
| **contrabando**: | Una base de datos es un conjunto de datos pertenecientes a un mismo contexto y almacenados sistemáticamente para su posterior uso. |
| **elusión**: | Cuando a un hecho imponible gravado se le da la apariencia de otro con la finalidad de disminuir o evitar el pago de tributos al fisco. |
| **evasión**: | Toda acción en contra de la ley del contribuyente –en contra de la ley– a fin |
| | de reducir, total o parcialmente, su carga tributaria y por ende la eliminación o disminución de su tributo. |
| **secreto fiscal**: | En el caso argentino el secreto fiscal se encuentra regulado por la la Ley de la Nación Nro. 11683[12], la cual dispone que: |
| | *«Las declaraciones juradas, manifestaciones e informes que los responsables o terceros presentan a la AFIP, y los juicios de demanda contenciosa en cuanto consignen aquellas informaciones, son secretos. Los magistrados, funcionarios, empleados judiciales o dependientes de la AFIP, están obligados a mantener el más absoluto secreto de todo lo que llegue a su conocimiento en el desempeño de sus funciones sin poder comunicarlo a persona alguna, ni aún a solicitud del interesado, salvo a sus superiores jerárquicos. Los terceros que divulguen o reproduzcan dichas informaciones incurrirán en la pena prevista por el artículo 157 del C.P., para quienes divulgaren actuaciones o procedimientos que por la ley deben quedar secretos».* |
| **trabajo no registrado**: | Existencia de empleo en relación de dependencia donde el empleador no comunica a la AFIP la existencia de trabajadores a su cargo. |

## Glosario de atributos-variables

|   |   |
|---|---|
| «accionista»: | El contribuyente posee en su acervo patrimonial acciones y/o cuotas partes de una sociedad determinada. |
| «accmayorit»: | El contribuyente posee en su acervo patrimonial una porción mayoritaria de acciones y/o cuotas partes que le permitiera controlar por sí solo una sociedad determinada. |
| «apellido»: | Apellido del contribuyente –persona física–. |
| «apellmater»: | Apellido materno del contribuyente –persona física–. |
| «asinpresdj»: | Períodos anuales, desde el inicio de actividad del contribuyente, que el mismo no presenta declaración jurada alguna. |
| «blanque-morat»: | El contribuyente decidió entrar en un régimen de moratoria y/o blanqueo de capitales. |
| «cantcau: | Existencia de causas judiciales iniciadas en contra del contribuyente. |
| categmonot»: | En caso de ser monotributista, categoría en la cual el contribuyente se encuentra inscripto. |
| «contribsan»: | Contribuyente sancionado, entendido como todo aquel que posea una o más sanciones y/o incumplimiento fiscal de algún tipo en el período fiscal en cuestión. |
| «CUIT»: | Clave única de identificación tributaria, clave única e irrepetible de cada contribuyente. |
| «decjurada»: | Si el contribuyente presenta o no declaración jurada. |
| «directivosoc»: | Entre sus actividades, el contribuyente es directivo de una sociedad determinada. |
| «donayoacred»: | El contribuyente efectuó donaciones y/o otorgó créditos a otros |

| | contribuyentes. |
|---:|:---|
| «estado»: | Estado activo o pasivo del contribuyente frente al fisco. |
| «fechanac»: | Fecha de nacimiento del contribuyente. |
| «liquidez»: | Proporción de liquidez patrimonial del contribuyente respecto el acervo patrimonial declarado, según su liquidez relativa esta puede ser Alta, Media o Baja. |
| «nroemple»: | Número de empleados en relación de dependencia y declarados por parte del contribuyente. |
| «nrodenuncias»: | Número de denuncias registradas contra el contribuyente. |
| «perfil»: | Perfil del contribuyente en relación a la actividad ejercida por el mismo. |
| «razonsocial»: | Razón social o nombre de la persona jurídica en cuestión. |
| «rellab-condicionIVA»: | Relación laboral, régimen de tributación y condición del contribuyente frente al IVA. |
| «siper»: | Acrónimo correspondiente a Sistema de Perfil de Riesgo del contribuyente. Metodología (SIPER) utilizada por el organismo para clasificar el riesgo del contribuyente según su incumplimiento. El mismo consta de cinco categorías alfabéticas que van desde «A» (menor riesgo) a «E» (mayor riesgo). |
| «superpodom»: | Superposición de domicilio entre dos o más contribuyentes, en el domicilio del contribuyente en cuestión. |
| «supdompfis»: | Superposición de domicilio entre el contribuyente y al menos otro contribuyente en la forma de persona física. |
| «supdompjud»: | Superposición de domicilio entre el contribuyente y al menos otro contribuyente en la forma de persona jurídica. |
| «tipopersona»: | Tipo de persona del contribuyente, ya sea una persona física o una jurídica. |

**RESUMEN**

Introduciendo elementos de ingeniería de explotación de la información al análisis tributario, a partir de herramental y conceptos computacionales avanzados de inteligencia artificial, en un enfoque epistemológico pragmático, se aborda la problemática del delito contra la hacienda pública. Mediante una aproximación empírica hipotética para un caso simulado, se aplican algoritmos de inducción, redes neuronales y redes bayesianas para determinar la factibilidad de su aplicación heurística en la administración pública fiscal. Distintas estrategias son exploradas para facilitar la labor local y regional del investigador tributario federal; esta vez, a partir de un enfoque computacionalmente limitado, pero igualmente eficaz para el experto en ciencias económicas avocado a la tarea de la investigación tributaria artesanal.

**Clasificación JEL:** C81; D80; H26; H83; H87
**Clasificación ACM:** H.1.1; I.2.1; I.2.4; I.5.0; K.4.1

**RESUMO**

Apresentando a introdução de elementos de exploração de informações para análise fiscal, por meio de software de mineração de dados e conceitos avançados computacionais de inteligência artificial, foi abordado o problema do crime de sonegador fiscal contra o patrimônio público. Através de uma abordagem empírica a partir de um caso hipotético de uso, os algoritmos de indução, redes neurais e redes bayesianas são aplicados para determinar a viabilidade de sua aplicação heurística pelo administrador público tributário. Diferentes estratégias são exploradas para facilitar o trabalho dos inspectores tributários federais locais e regionais, tendo em conta as suas capacidades computacionais limitados, mas igualmente eficaz para aqueles cientista social comprometido com a investigação fiscal.

**Classificação JEL:** C81; D80; H26; H83; H87
**Classificação ACM:** H.1.1; I.2.1; I.2.4; I.5.0; K.4.1